\documentclass{article}

\usepackage{arxiv}

\usepackage[utf8]{inputenc} 
\usepackage[T1]{fontenc}    
\usepackage{hyperref}       
\usepackage{url}            
\usepackage{booktabs}       
\usepackage{amsfonts}       
\usepackage{nicefrac}       
\usepackage{microtype}      

\usepackage{graphicx}
\usepackage{multirow}
\usepackage{amsmath}
\usepackage[htt]{hyphenat}
\usepackage{mathtools}
\usepackage{wrapfig}

\usepackage{subfig}
\usepackage[ruled,vlined]{algorithm2e}
\usepackage{import}

\newcommand{\X}{\mathcal{X}}

\renewcommand{\vec}[1]{\boldsymbol{#1}}     

\newcommand{\R}{\mathbb{R}}
\DeclareMathOperator{\E}{E}
\DeclareMathOperator{\Pro}{P}

\newcommand{\mymulticol}[1]{\multicolumn{2}{c|}{#1}}

\title{Image Inpainting Using Wasserstein Generative Adversarial Imputation Network}

\author{

  Daniel Vašata \\
  Faculty of Information Technology\\
  Czech Technical University in Prague\\
  Prague, Czech Republic\\
  \texttt{daniel.vasata@fit.cvut.cz} \\
  \And
  Tomáš Halama \\
  Faculty of Information Technology\\
  Czech Technical University in Prague\\
  Prague, Czech Republic\\
  \texttt{halamto2@fit.cvut.cz} \\
  \And
  Magda Friedjungová \\
  Faculty of Information Technology\\
  Czech Technical University in Prague\\
  Prague, Czech Republic\\
  \texttt{magda.friedjungova@fit.cvut.cz} \\
}

\date{}

\renewcommand{\undertitle}{A preprint of paper accepted to ICANN 2021.}

\begin{document}
\maketitle

\begin{abstract}
Image inpainting is one of the important tasks in computer vision which focuses on the reconstruction of missing regions in an image.
The aim of this paper is to introduce an image inpainting model based on Wasserstein Generative Adversarial Imputation Network.
The generator network of the model uses building blocks of convolutional layers with different dilation rates, together with
skip connections that help the model reproduce fine details of the output.
This combination yields a universal imputation model that is able to handle various scenarios of missingness with sufficient quality.
To show this experimentally, the model is simultaneously trained to deal with three scenarios given by missing pixels at random, missing various smaller square regions, and one missing square placed in the center of the image.
It turns out that our model achieves high-quality inpainting results on all scenarios.
Performance is evaluated using peak signal-to-noise ratio and structural similarity index on two real-world benchmark datasets, CelebA faces and Paris StreetView.
The results of our model are compared to biharmonic imputation and to some of the other state-of-the-art image inpainting methods.
\end{abstract}

\keywords{Imputation Methods  \and Missing Data \and Image Inpainting \and Generative Models \and Wasserstein GAIN \and Wasserstein GAN}

\section{Introduction}
\label{sec:intro}
In computer vision, one of the most important tasks being solved is image inpainting, also known as image completion,
which aims to restore missing pixels in a damaged image.
The aim is to estimate and impute the pixel information in missing locations
based on the context from non-missing parts of the image.
Since locations of missingness can appear in many ways such as random
noise or entire connected regions of various size and shape,
it may not be easy to have a universal model that can handle most of these scenarios.
Image inpainting can also be used for replacing unwanted by a realistically looking output.

Conventional approaches understand pixel imputation as a smooth function
extension problem, see e.g. \cite{bertalmio00,ballester01,simakov08,damelin18}.
These methods work well for cases where image corruption is minor or
straightforward to fill in, but not so well for cases with more significant damage,
failing to produce reasonable or plausible outcomes \cite{pathak16}.
Recently, the most successful methods (e.g. \cite{yu18,zheng19,shin20,jam20})
combine convolutional neural networks and generative adversarial networks which
yield improvements such as higher sharpness, matching colours
and general shapes of imputed objects in missing regions.
Typically these models have the common advantage that one does
not need to know which pixels are missing in advance.
However, the most successful ones are often of high-complexity and with complicated loss functions often based on pretrained networks
for visual classification.

The aim of this work is to address image inpainting task using
Wasserstein Generative Adversarial Imputation Network (WGAIN) that was recently
introduced by the authors in \cite{friedjungova20} as a general imputation model.
It is a generative imputation model which, for non-visual imputation tasks, performs comparatively to other state-of-the-art methods.
It beneficially incorporates the Wasserstein metric to adversarial
training which does not suffer from vanishing gradients.

For the image inpainting domain one needs to adjust the model for the
scenario of image data, namely make use of convolutional layers.
In our WGAIN model, we adopt the architecture from \cite{jam20} and extend it by using building blocks composed from
parallel convolutional layers with multiple dilation rates.
This leads to different sizes of the layers' receptive fields which improves the ability of the model to focus on both the local and global structure
of the image hence obtaining universality in terms of variable missing pixel regions. Moreover we use skip connections
allowing the model to propagate high resolution features in the hourglass network topology of the generator in a sandwich like way which
helps the model reproduce the fine details.

Our aim is to research the ability of our WGAIN model to perform well even without the highly complicated pre-trained elements.
We experimentally show that our model is able to perform well in three different scenarios of missingness when trained for all of them at once.
These scenarios are given by missing pixels at random,
missing various smaller square regions, and one missing square placed in the center of the image.
Hence the model is able to react properly on large missing areas as well as on many missing small areas simultaneously.
This shows the universality of the proposed WGAIN model.
The performance is evaluated using peak signal-to-noise ratio (PSNR) and structural similarity index (SSIM).
The results are compared to conventional methods of inpainting by biharmonic functions used e.g. in \cite{damelin18,amrani17,chen14,barnum17}.
We also discuss the comparison to other state-of-the art methods \cite{shin20,jam20,hua18,yu18,zheng19,pathak16} where possible.

\section{Related Work}
\label{sec:related}
Most conventional methods such as \cite{simakov08,ballester01,efros01,bertalmio00,telea2004image,damelin18}
used to perform computer-aided inpainting rely on local features such as colours and textures,
but they fail to consider the global semantics of the image.
These methods work well for cases where image corruption is minor or scattered across the image in small regions,
but not so well for cases with more significant regions to fill, failing to produce reasonable or plausible outcomes \cite{pathak16}.

A significant number of state-of-the-art methods use deep generative neural networks with very promising results.
One of the ways of creating globally well-organized and coherent images is by introducing
a second neural network, an adversary, that tries to decide whether the produced results look artificial or genuine.
The original generating network can learn to produce results that are much less likely
to be discarded as artificial using information from this adversary network.
Such networks are called generator and discriminator.
This type of architecture is called generative adversarial network (GAN) \cite{goodfellow14}.

Let us briefly mention some of state-of-the-art methods.
A very inspiring work handling inpainting using deep neural networks
with an adversary discriminative network is Context Encoders (CE) \cite{pathak16}.
Based on the autoencoder architecture and using only convolutional layers,
they achieved superior results in a semantic inpainting task.
In \cite{yu18} introduced contextual attention layer enables distant areas of the image to influence each other.
When combined with two discriminating losses, one for determining whether
the entirety of the resulting image is real-looking and one only for the generated patch,
the work achieved more plausible results than other methods in a human evaluated test.
Hui et al. in \cite{hui20} mitigated the problem of blurred outputs
using a one-stage model called dense multi-scale fusion network (DMFN), which utilizes dense combinations of dilated
convolutions to obtain larger and more effective receptive fields.
They designed a novel self-guided regression loss for
concentrating on uncertain areas and enhancing semantic details.
In \cite{zheng19} presented network contains reconstructive and generative parts, both represented by GANs,
and a new short+long term attention layer improving appearance consistency. This network is able to generate multi-modal results.
The PiiGAN \cite{cai20} based on \cite{yu18} also adopted the idea of producing multiple reasonable result. 
The recently proposed Symmetric Skip Connection Wasserstein Generative Adversarial Network \cite{jam20} contains encoder-decoder with convolutional blocks, linked by skip connections, together with a Wasserstein-Perceptual loss function to preserve colour and maintain realism on a reconstructed image.
PEPSI and Diet-PEPSI \cite{shin20} are another recent very successful GAN-based models incorporating parallel extended-decoder path for semantic
inpainting, which aims at reducing the number of convolution operations as well as improving the inpainting performance.
\section{Wasserstein Generative Imputation Network}
\label{sec:model}
Here we introduce the WGAIN following \cite{friedjungova20} closely.
Let us denote by $\X = \R^{m, n, 3}$ the space of all possible images of size
$m\times n$ and three color channels (RGB) and let $\vec X$
be a random element of $\X$ whose distribution is denoted by $\Pro(\vec X)$.
The identification of missing/damaged pixels is stored in a mask boolean matrix
$\vec{M} \in \{0,1\}^{m,n}$, where:
\[
\vec{M}_{i,j} = \begin{dcases*}
1, \text{\: if $ij$th pixel of } \vec{X} \text{ is valid,} \\
0, \text{\: if $ij$th pixel of } \vec{X} \text{ is missing.}
\end{dcases*}
\]
The distribution of $\vec M$ corresponds to the distribution of missingness in the data.
Let us further denote by $\tilde{\vec X}$ the image $\vec X$ having zeros in
place of missing pixels given by
\[
 \tilde{\vec X} = \vec X \odot \vec M,
\]
where $\odot$ denotes element-wise multiplication performed along all three color channels.

The next step is to prepare the input that can be used to replace the missing pixels in $\tilde{\vec X}$
by random values drawn independently from the normal distribution.
Formally, let $\vec Z \in  \R^{m, n, 3}$ be a random tensor with independent and identically distributed
components having normal distribution $\mathcal{N}(0,\sigma^2)$ with variance $\sigma^2$ and define
\[
 \tilde{\vec Z} = \vec Z \odot (1 - \vec M).
\]

To impute missing pixels in $\tilde{\vec X}$ based on the
information from non-missing pixels, we want the model to learn
the conditional distribution
$\Pro(\vec X | \tilde{\vec X}, \vec M )$ of $\vec X$ given
$\tilde{\vec X}$ and $\vec M$.

The generator $g$ of the WGAIN model is a mapping $g: \X \times \X \times \{0,1\}^{m,n} \to \X$
represented by a deep convolutional network that is fed by $\tilde{\vec X}$, $\tilde{\vec Z}$, and by $\vec M$.
It produces a new random image $g(\tilde{\vec X}, \tilde{\vec Z}, \vec M)$
corresponding to $\tilde{\vec X}$ with all pixels imputed.
The final image where only the missing pixels are imputed is then given by
\[
 \hat{\vec X}_{\vec Z} = g(\tilde{\vec X}, \tilde{\vec Z}, \vec M) \odot (1 - \vec M) + \tilde{\vec X} \odot \vec M
\]
and it is a random image whose conditional distribution
$\Pro(\hat{\vec X}_{\vec Z}| \tilde{\vec X}, \vec M)$
is given by the distribution $\Pro(\vec Z)$ of $\vec Z$ and should be
as close as possible to
$\Pro(\vec X | \tilde{\vec X}, \vec M)$.

The critic part $f$ of the WGAIN model is a Lipschitz mapping
$f: \X \times \{0,1\}^{m,n} \to \R$ represented by a deep convolutional network with norm restricted weights
and fed by images and masks trained to maximize
\[
 \E_{\vec X \sim \Pro(\vec X), \vec M \sim \Pro(\vec M)}\big(f(\vec X, \vec M)
 - \E_{\vec Z \sim \Pro(\vec Z)} f(\hat{\vec X}_{\vec Z}, \vec M)\big)
\]
which is estimated by sample means from mini-batches.
This corresponds to the estimate of the expectation with respect to $\vec M$ and $\vec X$
of the Earth-Mover's or Wasserstein distance~\cite{villani08,rubner97}
between the two conditional distributions
$\Pro(\hat{\vec X}_{\vec Z}| \tilde{\vec X}, \vec M)$
and $\Pro(\vec X | \tilde{\vec X}, \vec M)$.

\subsection{Training}
The critic $f$ is used in adversarial training of both the generator $g$ and the critic itself.
There the generator and the critic play an iterative two-player minimax game
where the critic wants to recognize the imputed values from the real ones and
the goal of the generator is to trick the critic so it cannot recognize them.
Moreover, the generator's output is tightened to the correct image by the absolute error loss function $\mathcal {L}_{\text{MAE}}$.

Therefore, there are two objective functions to minimize. The first corresponds to training of the
critic given by
\[
 J(f) = \E_{\vec X \sim \Pro(\vec X), \vec M \sim \Pro(\vec M)} \lambda_{f} \Big(f(\vec X, \vec M)
 - \E_{\vec Z \sim \Pro(\vec Z)} f(\hat{\vec X}_{\vec Z}, \vec M)\Big),
\]
where the weight $\lambda_{f}$ enables one to increase or decrease the influence of the corresponding gradient.
Second is the objective for the generator,
\[
 J(g) = \E_{\vec X \sim \Pro(\vec X), \vec Z \sim \Pro(\vec Z), \vec M \sim \Pro(\vec M)}\Big(- \lambda_g f(\hat{\vec X}_{\vec Z}, \vec M)
 + \lambda_{\text{MAE}} \mathcal {L}_{\text{MAE}}(\hat{\vec X}_{\vec Z}, \vec X)\Big),
\]
where $\lambda_g$ and $\lambda_{\text{MAE}}$ are weights enabling one to strengthen or weaken the influence of the absolute error loss function.

The pseudo-code of the WGAIN training is given in Algorithm~\ref{alg:WGAIN}.
The values of the objective functions are estimated from mini-batches.
The optimization is done via alternating gradient descent, where the
first step is updating the critic $f$ and the second step is
updating the generator $g$. Hence, when perfectly trained, the discriminator gives negative values for cases with imputed features and positive values for cases with true features. On the other hand, the generator entering the critic will be pushed to obtain large positive values of the critic as it gives to real values.

\begin{algorithm}[t]
\SetAlgoLined
\KwIn{$\alpha$ - the learning rate; $w_{\max}$ - maximal norm of critic weights used in clipping; $m$ - the mini-batch size;
$\lambda_{f}, \lambda_g, \lambda_{\text{MAE}}$ - weights of the objectives}
 Draw $m$ samples $\{\vec x_j\}_{j=1}^{m}$ from the dataset\;
 Draw $m$ samples $\{\vec m_j\}_{j=1}^{m}$ from the mask distribution\;
 Draw $m$ samples $\{\vec z_j\}_{j=1}^{m}$ from the normal distribution of $\vec Z$\;
 \While{not converged}{
 $\tilde{\vec x}_{\vec z_j} \leftarrow \vec z_j \odot (1 - \vec m_j) + \vec x_j \odot \vec m_j$\;
 $\hat{\vec x}_{\vec z_j} \leftarrow g(\tilde{\vec x}_{\vec z_j}, \vec m_j) \odot (1 - \vec m_j) + \vec x_j \odot \vec m_j$\;

 \vspace{1ex}
 Update weights $\vec w$ of $f$ using Adam optimizer with learning rate $\alpha$ and gradient

 $\nabla J(f) = \lambda_{f}\nabla\left[\frac{1}{m}\sum_{j=1}^m f\big(\hat{\vec x}_{\vec z_j}, \vec m_j\big)
  - \frac{1}{m}\sum_{j=1}^m f\big(\vec x_j, \vec m_j\big)\right]$\;

 \vspace{1ex}
 Clip the norm of $\vec w$ by $w_{\max}$\;
 \vspace{1ex}

 Update weights of $g$ using Adam optimizer with learning rate $\alpha$ and gradient

 $\nabla J(g) = \nabla\left[-\lambda_{g}\frac{1}{m}\sum_{j=1}^m f\big(\hat{\vec x}_{\vec z_j}, \vec m_j\big)
  + \lambda_{\text{MAE}}\frac{1}{m}\sum_{j=1}^m \lVert \hat{\vec x}_{\vec z_j} - \vec x_j\rVert^2\right]$\;
 }
 \caption{WGAIN training pseudo-code.}
 \label{alg:WGAIN}
\end{algorithm}

\subsection{Architecture of networks}
\begin{figure}
  \centering
  \includegraphics[width=1\textwidth]{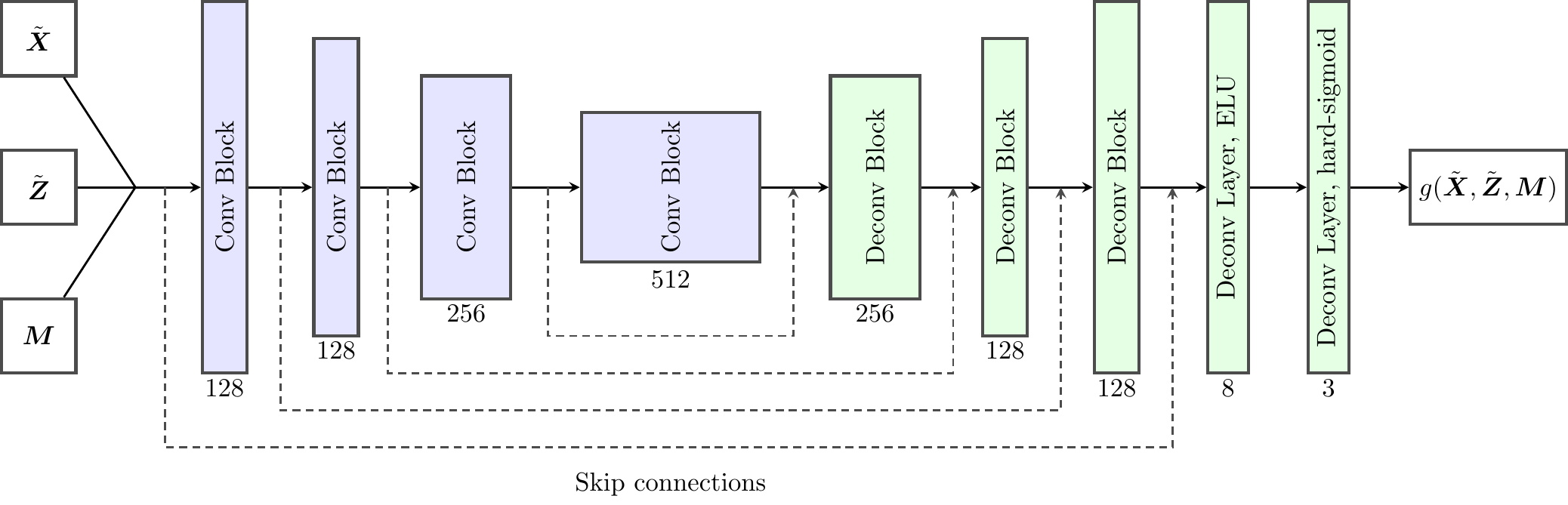}
  \caption{The architecture of the generator.}
  \label{fig:gen_architecture}
\end{figure}

Both the generator and the critic networks are based on convolutional layers.
The architecture of the generator $g$, as shown in Figure \ref{fig:gen_architecture},
is composed of building blocks of convolutional or deconvolutional layers with different dilation rates.
Those building blocks are then combined in the encoder-decoder bottleneck topology with sandwich like skip connections as introduced in \cite{jam20}.

The skip connections allow the model to propagate high resolution features from layers of the encoder into layers of the decoder (in reverse order)
which helps the model transfer the fine details in every depth better.
The first skip connection is fed by the concatenation of the network's input ($\tilde{\vec X}, \tilde{\vec Z}, \vec M$).
The subsequent ones by the outputs of the encoder's blocks.

The building blocks are composed of three parallel convolutional (for encoder) or deconvolutional (for decoder) layers with the same kernel size of $5 \times 5$
but with different dilation rates $0, 2, 5$ corresponding to different sizes of the layer's receptive field \cite{yu16dilated}.
The layers use padding and no strides so that the same dimension of the output is guaranteed.
The numbers of channels for the three layers
are of the form $(n/2, n/4, n/4)$ with increasing numbers in the encoder as $n = 128, 128, 256, 512$ and decreasing in the decoder as $n = 256, 128, 128$.
All three layers of the block have ELU activation functions and are concatenated into a single output.
In the case of the encoder the output goes into the outgoing skip connection and also into the next block.
If the next block belongs to encoder the max-pooling of pool size $2 \times 2$ is applied before entering it.
In the case of the decoder the input into the block is given by a concatenation of the previous block output and the incoming skip connection.
The output of the decoder's block is followed by an up-sampling operation of factor $2\times 2$.

The final block of the decoder is not up-sampled but only concatenated with the first skip connection and fed into the one other deconvolutional layer with $8$ channels,
kernel size of $3\times 3$, and ELU activation which is then followed by the last deconvolutional layer with $3$ channels, kernel size of $3 \times 3$, and hard-sigmoid activation function,
defined by
\[
  h(x) =
  \begin{cases}
    0 & \text{for } x < -2.5,\\
    0.2 x + 0.5 & \text{for } x \in [-2.5, 2.5],\\
    1 & \text{for } x > 2.5,
  \end{cases}
\]
that is responsible for collection of the final output.

The critic $f$ has a simple funnel topology with $5$ convolutional layers with kernel size of $5\times 5$, $2$ strides, and channel numbers $64, 128, 256, 256, 512$.
The layers have Leaky ReLU activation function.
The final output is produced by a single neuron connected to the flattened output of the last convolutional layer with linear activation.
The norm restriction needed for the Lipschitz property of the critic
is achieved by clipping the $L_2$ norm of each layer weights tensor to $1$.

\section{Experiments}
\label{sec:experiments}
The experiments were performed on two benchmark datasets:
Paris StreetView \cite{pathak16} and CelebA faces \cite{liu2015faceattributes}.
For the CelebA faces dataset the aligned and cropped variant which has faces aligned in the central position was used.
In the preprocessing step images from both datasets were cropped to be square shaped and have a common size of $128 \times 128$ pixels.


\subsection{Scenarios of missingness}
\label{subsec:masks}
In order to analyze the performance of the inpainting model we focus on three scenarios of missingness, i.e. on three probability distributions of the mask $\vec M$.
These three scenarios can be taken as representatives of three qualitatively different situations of how the missing pixels might be distributed across the image.
\begin{description}
  \item[Noise] corresponds to the situation when each pixel of the mask $\vec M$ is sampled independently
    on other pixels with a probability $p$ of having value $0$ which corresponds to the portion of missingness.
    In this scenario, we choose three different values of $p$ to simulate various damage portions. The simplest case is when $50\%$ of the pixels are dropped.
    The more severe damages are represented by $75\%$ and $95\%$.

    In the training phase, the values of $p$ for each sample are generated randomly with a uniform distribution in the interval $[0.5, 0.95]$.
  \item[Single square in the center] represents a demanding task
    with a large continuous region missing in the image,
    as there are no hints left inside the area.
    To test this scenario, we fixed $\vec M$ to represent a centered square of missing pixels. One side of the missing square is as long as half of the side of
    the original image,
    thus the missing portion is $25\%$.

    In the training phase the square is centered but its side is a randomly (uniformly) chosen integer in the interval $[\ell/2.5, \ell/1.6]$, where $\ell$ is the side of the original image.
  \item[Randomly located multiple squares] is a compromise between
    the previous two types of region mask. There are multiple smaller squares uniformly independently
    distributed across the image. The number of randomly located squares
    is fixed to $5$ and the squares have a fixed size of
    $31 \times 31$ pixels. Because of the overlapping it yields the final
    missing portion approximately equal to $25\%$.

    In the training phase the number of squares, their positions, and their sizes are chosen randomly. To be precise, we generate $30$ squares with
    lower left corners uniformly distributed in the 2D interval $[-2\ell, 3\ell]^2$ and with their sides uniformly distributed in the interval $[\ell/5, \ell/3]$.
    The final mask for the sample is then given by the intersection of those squares with the 2D interval $[1, \ell]^2$.
\end{description}

During the training phase the model learns all these scenarios at once. This means that each training sample randomly choses which scenario it belongs to and then
it generates the mask matrix as described above. In the evaluation phase each of these scenarios is evaluated separately.

\subsection{Implementation details}
\label{subsec:implementation}
We perform a global normalization on all channels of the images to set the intensity values of the pixels
in the range $[0,1]$.
The hyperparameters for the experiment were empirically set as $\lambda_f = 1$, $\lambda_g = 0.005$, and $\lambda_{\text{MAE}} = 1$.
The training procedure was optimized using Adam optimizer with learning rate $\alpha = 0.00005$.
The mini-batch size was $m = 32$.
The model for the Paris StreetView dataset was trained in $2000$ epochs and the model for the CelebA dataset in $200$ epochs.
This corresponds to a similar number of training steps and training time for
both datasets.

The source code of our experiments is available at Github repository\footnote{\url{https://github.com/vasatdan/wgain-inpaint}}.
We used the~\texttt{TensorFlow} library\footnote{\url{https://www.tensorflow.org}} running on a nVidia Tesla V100-PCIE-32GB.
It took approximately $3$ days to train each model. For the implementation of biharmonic function inpainting we used the \mbox{scikit-image}\footnote{\url{https://scikit-image.org/}} library.

\subsection{Results}
\label{sec:results}
\begin{figure}
  \centering
  \includegraphics[width=0.9\textwidth]{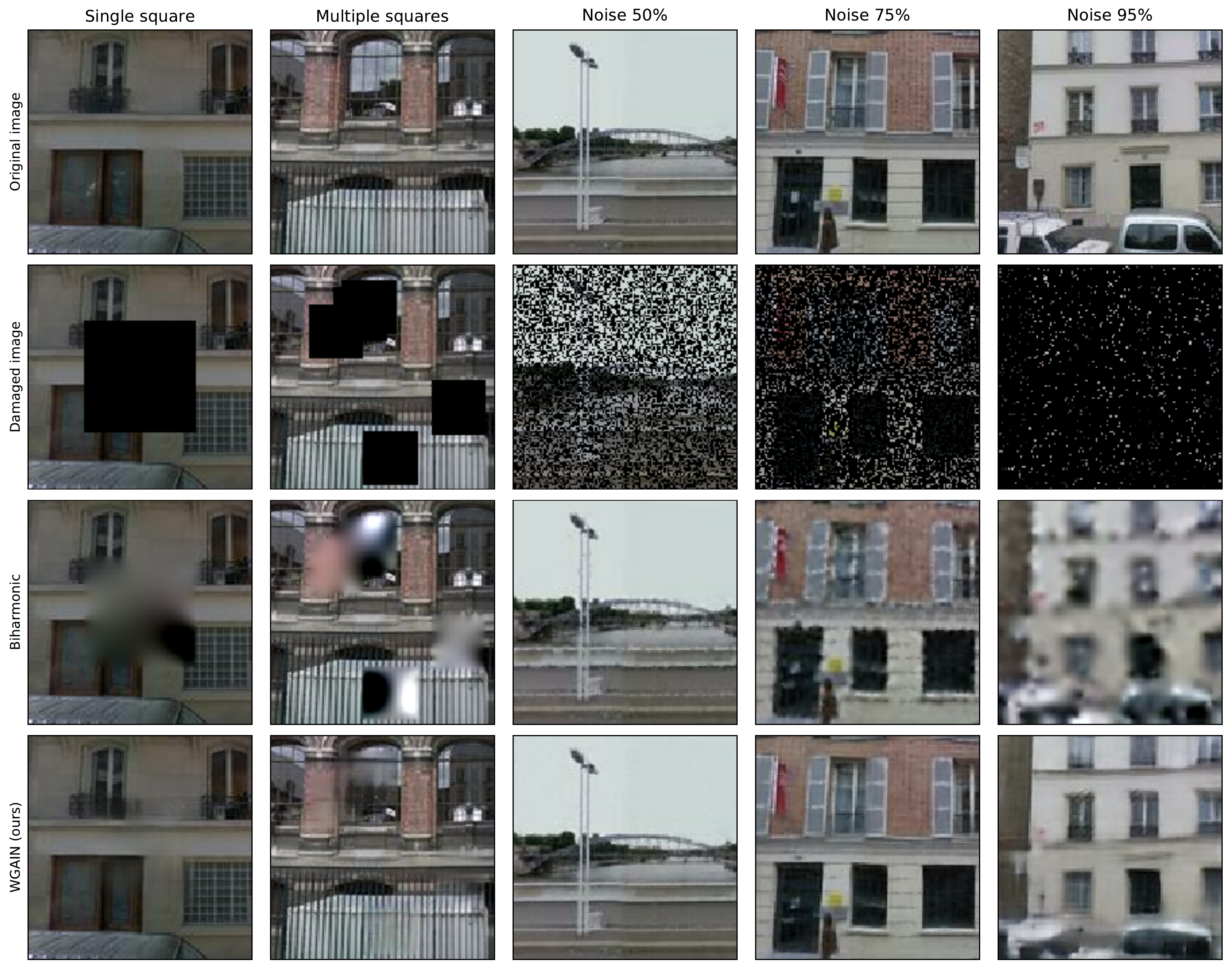}
  \caption{Demonstration of inpainting scenarios and results for Paris StreetView dataset.}
  \label{fig:damages1}
\end{figure}

\begin{figure}
\centering
\includegraphics[width=0.9\textwidth]{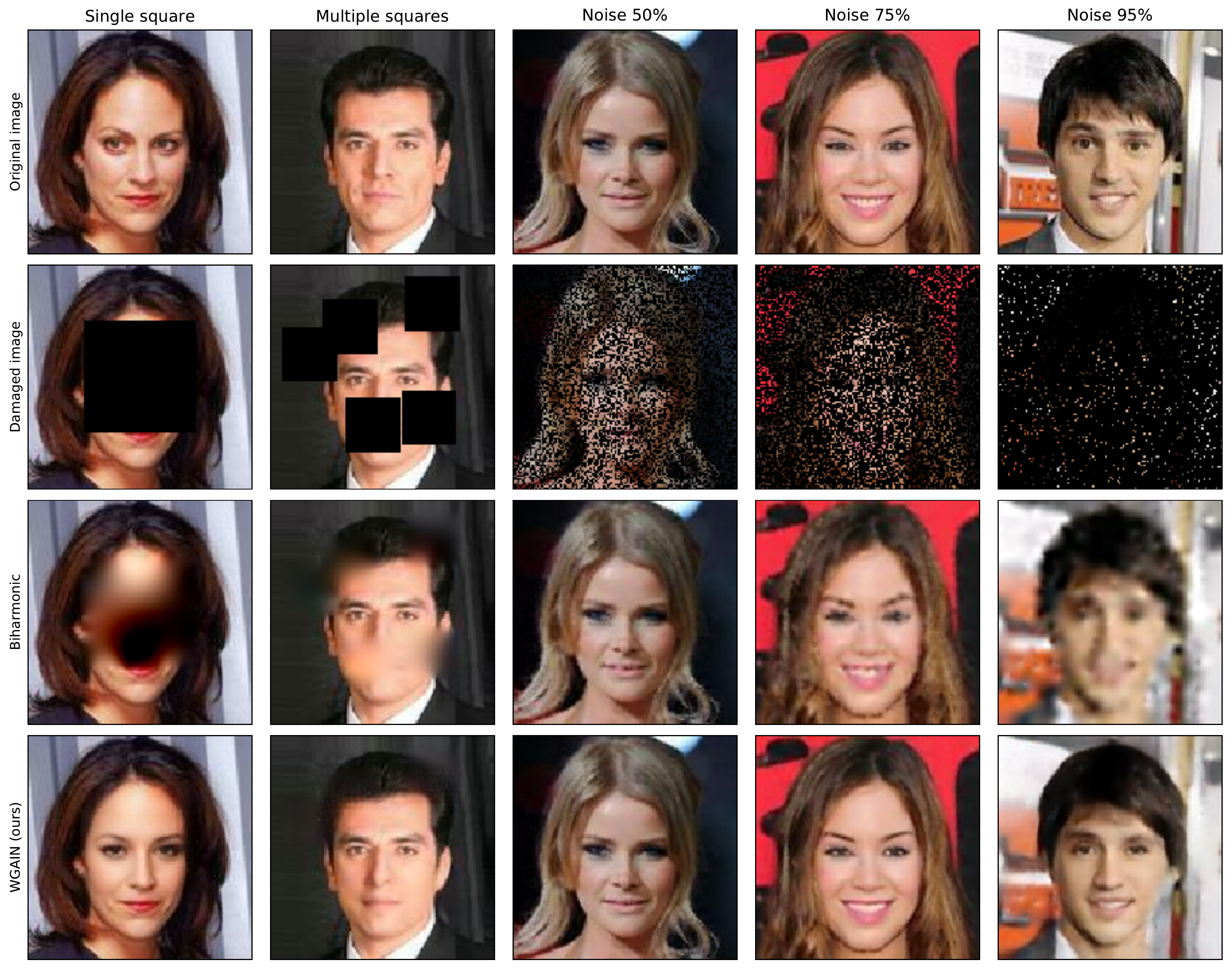}
\caption{Demonstration of inpainting scenarios and results for CelebA dataset.}
\label{fig:damages2}
\end{figure}

The examples of the experimental results are shown in Figures \ref{fig:damages1} and \ref{fig:damages2}.
Our model performs well for both datasets in all scenarios of missingness. Moreover, in all cases it visually outperforms
the results of inpainting by biharmonic functions. Interesting results can be observed in Figure \ref{fig:damages2} in the single centered square
scenario.
Here the inpainted face looks quite realistic but differs from the original image.
The person on the original image is looking to the left with eyes wide open whereas the face generated by our model is looking to the center with less open eyes.
We may say that the inpainting result is satisfactory since one is not able to determine this information from the non-missing part of the image.


As a quantitative evaluation the peak signal-to-noise ratio (PSNR) \cite{wang09}
and the structural similarity index measure (SSIM) \cite{wang04} were used.
Both metrics are common for image inpainting evaluation \cite{cai20,hui20,shin20,yu18}.
In Table \ref{tab:res2} the results are presented together with
biharmonic function inpainting results in the same setup.
In all evaluation scenarios the WGAIN outperformed biharmonic inpainting.

\begin{table}[h]
  \caption{Results on Paris StreetView and CelebA datasets.}
  \label{tab:res2}
  \centering
  \begin{tabular}{|c|c|c|c|c|c|c|c|c|c|c|c|c|c|c|c|c|c|}
    \hline

    \multicolumn{2}{|c|}{\multirow{2}{*}{Damage type}}  & \multicolumn{8}{c|}{Paris StreetView}  & \multicolumn{8}{c|}{CelebA}  \\ \cline{3-18}
    \multicolumn{2}{|c|}{} & \multicolumn{4}{c|}{WGAIN}  & \multicolumn{4}{c|}{Biharmonic} & \multicolumn{4}{c|}{WGAIN}  & \multicolumn{4}{c|}{Biharmonic} \\ \cline{3-18}
    \multicolumn{2}{|c|}{} & \multicolumn{2}{c|}{PSNR} & \multicolumn{2}{c|}{SSIM}  & \multicolumn{2}{c|}{PSNR} & \multicolumn{2}{c|}{SSIM} & \multicolumn{2}{c|}{PSNR} & \multicolumn{2}{c|}{SSIM}  & \multicolumn{2}{c|}{PSNR} & \multicolumn{2}{c|}{SSIM} \\ \hline
    \multicolumn{2}{|c|}{Singlesquare 25\%} & \mymulticol{\textbf{25.00}} & \mymulticol{\textbf{0.88}} & \mymulticol{21.12} & \mymulticol{0.85} & \mymulticol{\textbf{25.96}} & \mymulticol{\textbf{0.92}} & \mymulticol{17.94} & \mymulticol{0.83} \\ \hline
    \multicolumn{2}{|c|}{Multisquare 25\%}   & \mymulticol{\textbf{26.51}} & \mymulticol{\textbf{0.90}} & \mymulticol{22.67} & \mymulticol{0.86}  & \mymulticol{\textbf{26.75}} & \mymulticol{\textbf{0.93}} & \mymulticol{23.34} & \mymulticol{0.89}    \\ \hline
    \multirow{3}{*}{Noise} & 50\% & \mymulticol{\textbf{31.48}} & \mymulticol{\textbf{0.96}} & \mymulticol{30.11} & \mymulticol{0.95} & \mymulticol{\textbf{34.00}} & \mymulticol{\textbf{0.98}} & \mymulticol{33.37} & \mymulticol{0.98}        \\ \cline{2-18}
    & 75\%         & \mymulticol{\textbf{27.73}} & \mymulticol{\textbf{0.90}}& \mymulticol{25.90} & \mymulticol{0.87} & \mymulticol{\textbf{29.96}} & \mymulticol{\textbf{0.95}} &  \mymulticol{28.73} & \mymulticol{0.93}       \\ \cline{2-18}
    & 95\%     & \mymulticol{\textbf{22.72}} & \mymulticol{\textbf{0.74}} & \mymulticol{21.13} & \mymulticol{0.67}   & \mymulticol{\textbf{23.86}} & \mymulticol{\textbf{0.83}} & \mymulticol{22.52} & \mymulticol{0.79}      \\ \cline{2-18} \hline

  \end{tabular}
\end{table}

To be able to compare the results to other state of the art methods, we used the single square in the center scenario.
The values of the PSNR and SSIM measures for PiiGAN, DMFN, and CE compared to our method are summarized in Table \ref{tab:methods}.
It shows that on the Paris StreetView dataset the WGAIN outperforms CE and also the DMFN in SSIM with equal PSNR.
On the CelebA faces dataset in comparison to the DMFN our model has lower PSNR and higher SSIM. Both WGAIN and DMFN, however, are outperformed by the PiiGAN for this dataset.

To interpret this comparison correctly one should note that the results of the experiments presented for the other methods were often obtained with different
resolutions of images, for different target tasks, and some of them actually on different datasets - instead of the CelebA dataset, the CelebA-HQ dataset collected
from CelebA and post-processed (for details see \cite{karras17}) was used in both \cite{cai20,hui20}.
Especially the different target tasks are of high importance. The presented results for the competitive models are
obtained under the scenario where the corresponding imputation method
is trained on the same task where it is evaluated.
It means that the models are trained to impute the centered square of fixed size only.
On the other hand, our model is trained for all the scenarios of missingness together and performs quite well on all of them.
Hence, on one specific subtask, it might be outperformed by a specialized model trained for that subtask only.

\begin{table}
  \caption{Comparison of inpainting methods on the single square in the center scenario of missingness, where $25\%$ of pixels are missing. The values of PSNR and SSIM are taken from the papers cited in the table.
  Note that PiiGAN and DMFN used CelebA-HQ dataset, and that DMFN used images of size $256 \times 256$.}
  \label{tab:methods}
  \centering
  \begin{tabular}{|c|c|c|c|c|}
    \hline
    \multicolumn{1}{|c|}{\multirow{2}{*}{Method}}  & \multicolumn{2}{c|}{CelebA dataset}  & \multicolumn{2}{c|}{Paris StreetView dataset} \\ \cline{2-5}
    \multicolumn{1}{|c|}{} & PSNR & SSIM & PSNR &  SSIM \\
    \hline
    PiiGAN\cite{cai20} & 34.99 & 0.99 & - & - \\
    \hline
    DMFN\cite{hui20} & 26.50 & 0.89 & 25.00 & 0.86 \\
    \hline
    CE\cite{pathak16}  & - & - & 18.58 & - \\
    \hline
    WGAIN (ours) & 25.96 & 0.92 & 25.00 & 0.88\\
    \hline
  \end{tabular}
\end{table}

\section{Conclusion}
\label{sec:conclusion}
In this paper we present an image inpainting model based on Wasserstein Generative Adversarial Imputation Network
where the generator network uses convolutional building blocks and skip connections.
The combination of convolutional layers with different dilation rates enables each building block to focus on both the global (large range) and the local (small range)
structure of the input, and skip connections help the model reproduce fine details of the output.

This yields a universal imputation model that is able to handle various scenarios of missingness with sufficient quality.
We tested three scenarios given by missing pixels at random, missing various smaller square regions, and one missing square placed in the center of the image.
The model was trained simultaneously for all of the scenarios.
The performance was evaluated using peak signal-to-noise ratio and structural similarity index on two real-world benchmark datasets, CelebA faces and Paris StreetView.
The results were compared to biharmonic imputation and to three other state-of-the-art methods.
It turns out that our WGAIN image inpainting model achieves high-quality inpainting results which outperform the conventional inpainting by biharmonic functions
and is comparable to state-of-the-art method DMFN\cite{hui20}. The superiority of PiiGAN\cite{cai20} on the CelebA dataset compared to our model is assumed to be caused by focusing on only one scenario of missingness.
%

\section*{Acknowledgements}
This research has been supported by SGS grant No. SGS20/213/OHK3/3T/18, by GACR grant No. GA18-18080S,
and by the Student Summer Research Program 2020 of FIT CTU in Prague, Czech Republic.


%
%
\bibliography{arxiv}{} 
\bibliographystyle{unsrt}


\end{document}